\documentclass{article}

\usepackage{amsmath}
\usepackage{array}
\usepackage{authblk}
\usepackage{booktabs}
\usepackage[
  colorlinks=true,
  allcolors=blue,
  hyperfootnotes=false
]{hyperref}
\usepackage[english]{babel}
\usepackage{float}
\usepackage[symbol]{footmisc}
\usepackage{graphicx}
\usepackage[letterpaper,top=2cm,bottom=2cm,left=3cm,right=3cm,marginparwidth=1.75cm]{geometry}
\usepackage{multirow}
\usepackage{setspace}
\usepackage{tabularx}
\usepackage{xltabular}
\title{Curated endoscopic retrograde cholangiopancreatography images dataset}

{\author[1, \thanks{These authors contributed equally.}]{Alda João Andrade}}
{\author[2, \textsuperscript{\dag}]{Mónica Martins}}
\author[2,3]{André Ferreira}
\author[1]{Tarcísio Araújo}
\author[1,4,5, \thanks{These authors jointly supervised this work.}]{Luís Lopes}
\author[2, \textsuperscript{\ddag}]{Victor Alves}

\affil[1]{Department of Gastroenterology, Hospital Santa Luzia, ULS Alto Minho, Viana do Castelo, Portugal}
\affil[2]{Center Algoritmi / LASI, University of Minho, Braga, 4710-057,  Portugal}
\affil[3]{Institute for Artificial Intelligence in Medicine (IKIM), Essen University Hospital
(AöR), University of Duisburg-Essen, Essen, Germany}
\affil[4]{Life and Health Sciences Research Institute (ICVS), School of Medicine, University of Minho, Braga, Portugal}
\affil[5]{ICVS/3B’s - PT Government Associate Laboratory, Braga/Guimarães}

\begin{document}
\maketitle

\begin{abstract}
Endoscopic Retrograde Cholangiopancreatography (ERCP) is a key procedure in the diagnosis and treatment of biliary and pancreatic diseases. Artificial intelligence has been pointed as one solution to automatize diagnosis. However, public ERCP datasets are scarce, which limits the use of such approach.
Therefore, this study aims to help fill this gap by providing a large and curated dataset. The collection is composed of 19.018 raw images and 19.317 processed from 1.602 patients. 5.519 images are labeled, which provides a ready to use dataset. All images were manually inspected and annotated by two gastroenterologist with more than 5 years of experience and reviewed by another gastroenterologist with more than 20 years of experience, all with more than 400 ERCP procedures annually.
The utility and validity of the dataset is proven by a classification experiment.
This collection aims to provide or contribute for a benchmark in automatic ERCP analysis and diagnosis of biliary and pancreatic diseases.
\end{abstract}

\section{Background \& Summary}

Endoscopic Retrograde Cholangiopancreatography (ERCP) is a key procedure in the diagnosis and treatment of biliary and pancreatic diseases, yet publicly available imaging datasets documenting its fluoroscopic component are extremely limited. At the same time, recent reviews have highlighted a growing interest in applying Artificial Intelligence (AI) approaches to pancreaticobiliary endoscopy, including ERCP, for tasks such as the prediction of post-ERCP pancreatitis and the differentiation between malignant and benign bile-duct strictures \cite{chi2023risk, brenner2024development, vu2025deep}. Despite this increasing interest, this field is consistently described as being at an early stage of development, primarly due to methodological and data-related limitations \cite{goyal2021application,bharwad2025artificial}. Most existing studies are based on small, single-center datasets, which undermines reproducibility and limits the generalization of the proposed models \cite{jovanovic2014artificial,vu2025deep, ruffle2019artificial}.

Moreover, the lack of publicly available fluoroscopic ERCP image repositories hinders the development of standardized benchmarks for tasks such as image classification, lesion detection and procedural quality assessment. Therefore, this dataset was compiled to help address these limitations by providing a structured and anonymized collection of ERCP fluoroscopic images obtained during routine clinical practice at a tertiary care center. Through systematic selection and preprocessing, the dataset offers a resource that researchers can use to explore computational methods applicable to real-world ERCP imaging, supporting methodological studies in machine learning, computer vision, and interventional endoscopy.

The dataset is composed of fluoroscopy images acquired during ERCP procedures and includes clinically relevant categories such as biliary lithiasis, biliary leaks, benign strictures, malignant strictures, and normal findings. The images are made publicly available and are intended to complement existing endoscopic imaging resources \cite{borgli2020hyperkvasir,smedsrud2021kvasir,bravo2025gastrohun}.
The dataset may be used in studies involving image-based analysis of therapeutic endoscopy procedures, including the development and evaluation of automatic image classification methods.

Labeled and unlabeled images are present in the dataset, allowing its use in fully supervised learning approaches when considering only the labeled subset, as well as in semi-supervised or weakly supervised learning research that combines both labeled and unlabeled data. Furthermore, the collection can be used for transfer learning and data augmentation, facilitating the adaptation of existing models to fluoroscopic imaging and other medical imaging modalities.
Beyond research, the dataset can be used as an educational resource for trainees and specialists in gastroenterology and radiology, as well as for computer vision researchers, promoting interdisciplinary training in ERCP interpretation and AI applications in gastroenterology.

\section{Methods}

The DICOM files were retrospectively extracted from the clinical archive of a tertiary care hospital using the SECTRA PACS system, enabling the collection of ERCP images acquired between 2015 and 2025. All procedures were performed by gastroenterologists with more than five years of ERCP experience and an annual case volume exceeding 400 procedures, at a high-volume ERCP center performing over 750 procedures annually. Fluoroscopic images were acquired by radiology technicians present in the procedure room. All examinations were subsequently reviewed independently by a second gastroenterologist to ensure consistency and diagnostic reliability.  Retrieval from the PACS allowed for automated export of full image series, ensuring comprehensive coverage of the fluoroscopic examinations.
The dataset includes all fluoroscopic images acquired throughout the ERCP workflow, encompassing pre-cannulation images used for anatomical orientation and procedural planning, cholangiographic images obtained during contrast injection and therapeutic maneuvers, as well as post-procedural images documenting the final outcome of the intervention.
The images were acquired using different fluoroscopic systems, including the Ziehm Vision RFD 3D, and Philips PCR Eleva units, resulting in natural variability in image characteristics across the dataset.
Following extraction from the SECTRA PACS system, all DICOM files underwent a multi-stage Python-based processing pipeline designed to organize, extract, segment and anonymize the final dataset. The dataset is available in \url{https://doi.org/10.6084/m9.figshare.31079236}. The workflow comprised the steps described in the following sub-sections.

\subsection{Ethics statement*} 

The study was approved by the Board of Directors of Unidade Local de Saúde do Alto Minho (ULSAM). Approval ID: 243/CA-2025.

\subsection{DICOM extraction and creation of the primary dataset} 

All exported files were initially distributed across multiple ZIP archives. These were batch-unpacked, and the individual DICOM files processed sequentially.
For each DICOM file:

\begin{enumerate}
    \item The file was read using pydicom.dcmread.
    \item The pixel data were extracted and converted into a PNG image, saved in the "raw" directory.
    \item Relevant metadata were retrieved using standard DICOM tags:
    \begin{itemize}
        \item Patient ID (0010,0020)
        \item Patient Sex (0010,0040)
        \item Patient Birth Date (0010,0030)
        \item Content Date (0008,0023)
        \item Acquisition Time (0008,0032)
        \item Manufacturer's Model Name (0008,1090)
    \end{itemize}
    \item A corresponding entry was appended to an initial metadata CSV file.
\end{enumerate}

Raw images followed the naming convention:

\begin{equation*}
    \{patient\_id\}\_image\{N\}.png
\end{equation*}

where N is an incrementing index per patient.
This stage produced 19.018 PNG images corresponding to 1.602 patients and their metadata.

\subsection{Manual classification of images into S (single) and V (vertical multipanel)}

The raw images in the dataset were acquired using multiple fluoroscopic systems, each with distinct acquisition and export characteristics. Consequently, the dataset exhibited substantial variability in image structure: some systems produced isolated single-frame fluoroscopic images, whereas others generated multi-frame sequences exported as vertically stacked composite images. This heterogeneity in image morphology necessitated a manual classification step to ensure dataset standardization and to enable consistent downstream processing and analysis.

Following visual inspection, each image was assigned to one of two categories:
\begin{itemize}
    \item \textbf{S} – single-frame fluoroscopic image
    \item \textbf{V} – vertically stacked multi-frame composite image
\end{itemize}

This classification was recorded in the dataset's metadata file as a dedicated column named \textit{image\_type}, providing an explicit identifier for subsequent preprocessing and modeling tasks.

\subsection{Automated partitioning of multi-vertical (V) images}

\begin{figure}
    \centering
    \includegraphics[width=1\linewidth]{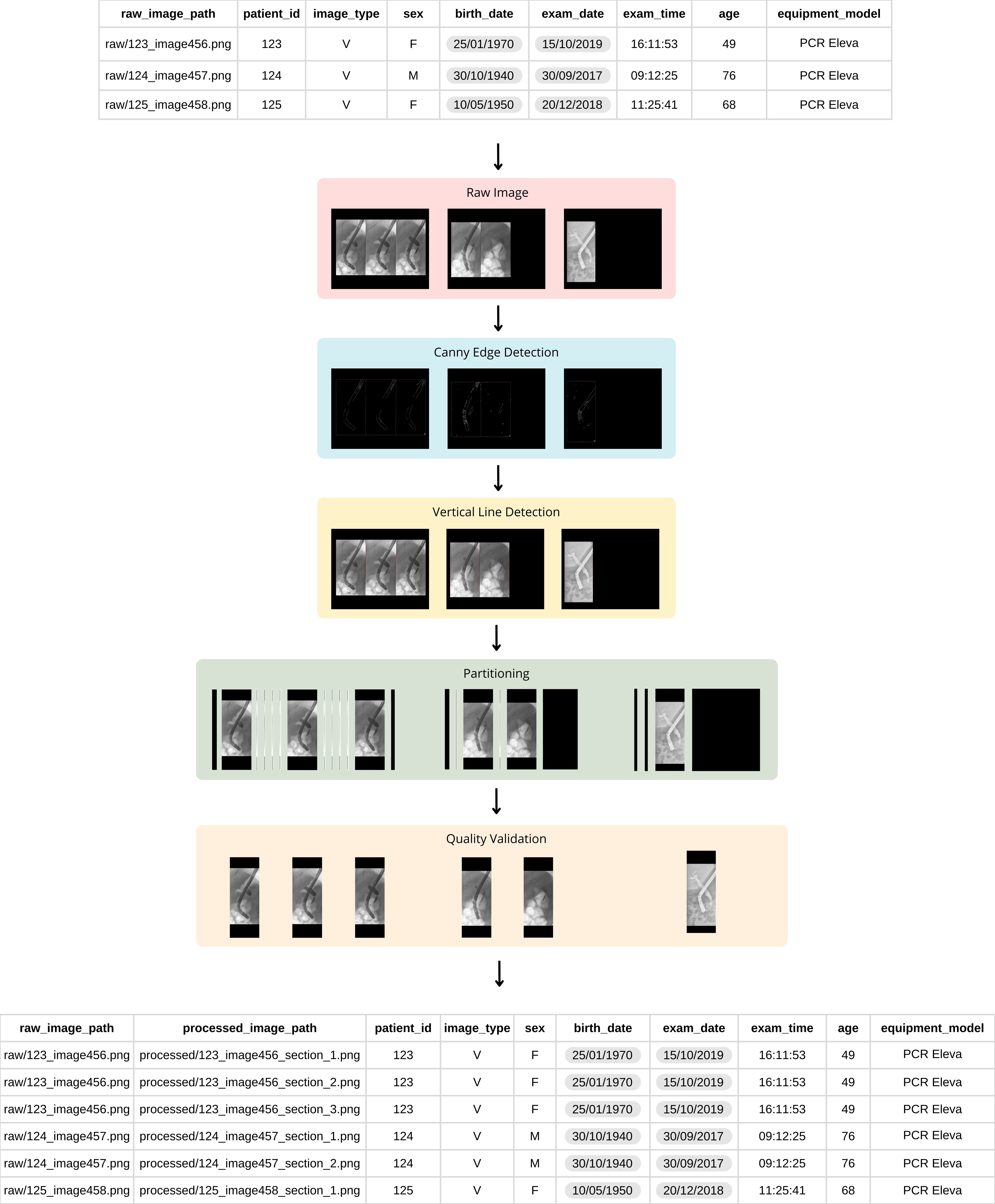}
    \caption{Data preprocessing pipeline.}
    \label{fig:preprocess_pipeline}
\end{figure}

Images classified as multi-frame vertical composite (V) frequently contained multiple fluoroscopic frames displayed side-by-side within a single composite DICOM image. To isolate these frames, an automated partitioning procedure was applied, as illustrated in Figure \ref{fig:preprocess_pipeline}.

First, each V image was processed using the Canny edge detector to enhance structural boundaries and facilitate vertical-line detection. This procedure increased the visibility of structural discontinuities and emphasized intensity transitions, which are characteristic of the vertical separators commonly generated by fluoroscopic acquisition systems.

Following edge detection, probabilistic Hough transform-based line detection was applied to identify candidate vertical lines. Only lines whose orientation deviated minimally from the vertical axis were retained, ensuring robustness against minor geometric distortions introduced by the fluoroscopic system. The detected lines were then sorted from left to right, and their x-coordinates were used as boundaries to partition each composite image into individual vertical segments.

After partitioning, each candidate section underwent a quality validation step designed to exclude invalid or unusable images. Segments were discarded if they exhibited any of the following characteristics:
\begin{itemize}
    \item Insufficient width (indicating partitioning errors or extremely narrow sections).
    \item Extremely low intensity across most of the image (indicating blank or near-blank frames).
    \item Complete saturation (fully white or fully black images).
    \item Insufficient contrast between minimum and maximum pixel intensities.
\end{itemize}

Only segments satisfying all quality criteria were retained. In addition to automated filtering, all cases were independently reviewed by a gastroenterologist and subsequently re-evaluated by a senior clinician to ensure the accuracy and reliability of the final dataset.

For each valid segment, a new entry was created in the metadata file and a corresponding PNG image was saved in the "processed" directory, following the convention:
\begin{equation*}
    \{patient\_id\}\_image\{N\}\_section\_\{k\}.png
\end{equation*}

As a result, a single "V" type image could generate multiple processed outputs, each representing an independent fluoroscopic frame. After this process, a total of 19.317 images were generated and stored in the "processed" folder, matching the number of entries recorded in the metadata CSV file.

To maintain consistency in the dataset structure, single-frame (S) images were also copied into the "processed" folder, although they did not undergo partitioning. This ensured that the "processed" directory contains a one-to-one correspondence with the metadata entries and provides a uniform source of input for downstream computational tasks.

\subsection{Dataset anonymization}

To ensure complete removal of identifiable information, all images of type "S" (both raw and processed), which contained patient identifiers, were anonymized using a standardized geometric transformation:
Central square crop, preserving only the largest possible centered square region.
Application of a circular mask, keeping only the circular region and setting the remaining area to black.
This transformation removes all bordering regions where patient identifiers or contextual screen information may appear. Images which type was "V" did not need any anonymization.
Finally, original patient IDs were replaced with a fully anonymized incremental index. A mapping table linking original to anonymized IDs was generated and stored separately.

\subsection{Image Annotation}

Following data anonymization and section-based partitioning, a subset of images was manually annotated into the categories "lithiasis", "biliary leaks", "malignant stricture", "benign stricture" and "normal". These labels were added to the metadata.csv file under the column "Label". Additionally, a “Keep” column was introduced, with values "keep" or "discard", to indicate whether each image adequately displayed the relevant diagnostic finding or whether it represented a non-informative frame, such as pre-contrast, calibration or otherwise non-diagnostic images. Images that were not manually annotated were automatically assigned the labels “Unlabelled” and “Discard”.

Annotations were performed by two gastrointestinal endoscopists. To ensure robustness and mitigate single-expert bias, every annotated image was subsequently reviewed by a third senior clinician and cross-checked against the endoscopist’s procedural report to verify consistency with the documented findings.

\section{Data Records}

All images were originally acquired using fluoroscopic systems during routine ERCP procedures.
Images were exported in DICOM format and reviewed by experienced endoscopists. 

\begin{figure}[H]
    \centering
\includegraphics[width=0.5\linewidth]{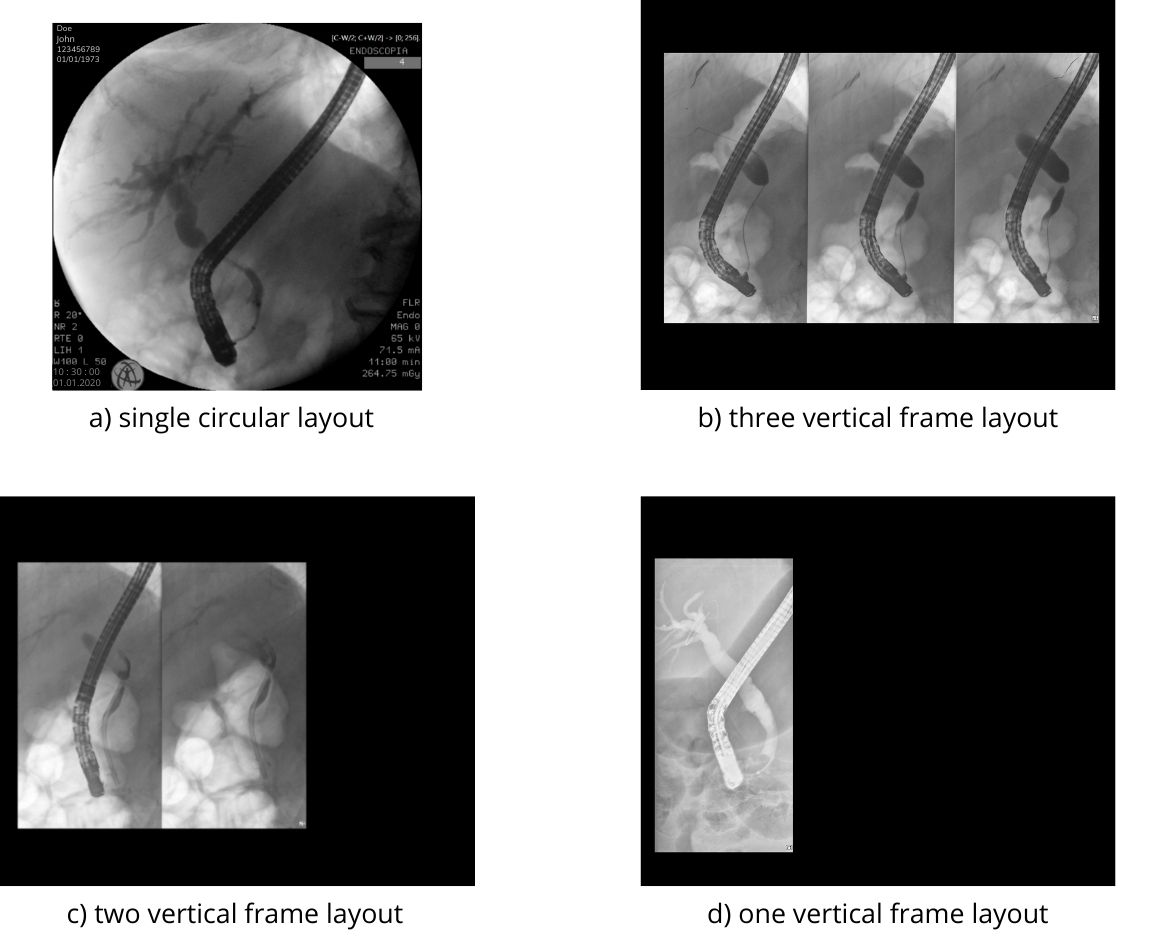}
    \caption{Samples from the dataset, showing the heterogeneity of the images.}
    \label{fig:examples}
\end{figure}

The dataset is divided into "raw" and "processed" folders. The "raw" data contains PNG images obtained directly from the original DICOM files, without almost any data processing, i.e., only anonymization was applied to the data. Figure \ref{fig:examples} presents some samples, highlighting the heterogeneity of the dataset. The "processed" data consists of cropped portions of the raw images. Each raw image, originally containing multiple sub-images, was partitioned into individual images, providing ready to apply data, allowing their direct use in downstream tasks, perhaps requiring only minimal adaptation. The pre-processing pipeline is detailed in Figure \ref{fig:preprocess_pipeline}.

\begin{figure}[H]
    \centering
\includegraphics[width=0.25\linewidth]{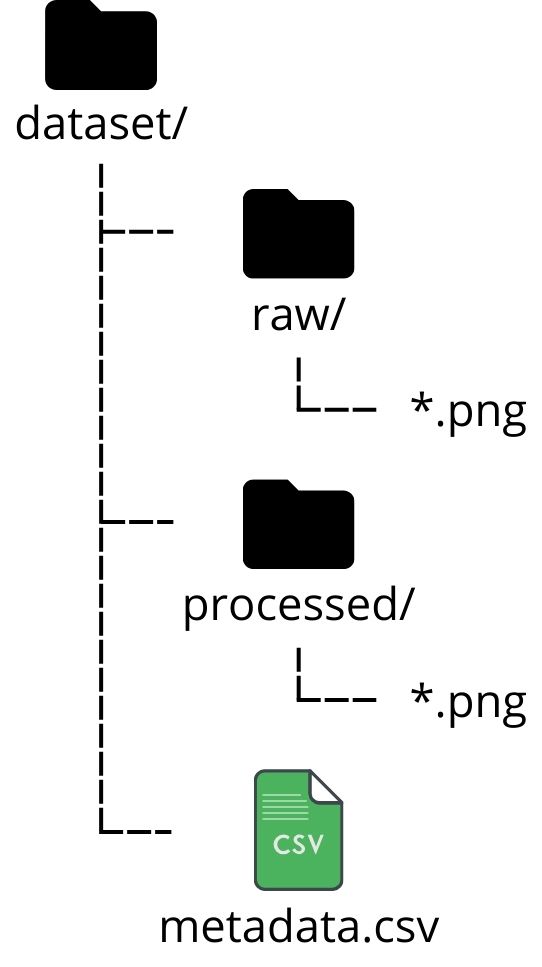}
    \caption{Schematic of the folders structure.}
    \label{fig:folder_structure}
\end{figure}

Figure \ref{fig:folder_structure} presents a schematic of the folders structure.
The "raw" folder contains 19.018 PNG images, while the "processed" folder contains 19.317 PNG images, reflecting the fact that a single raw image may yield multiple processed sections.

A CSV file, \textit{metadata.csv}, links each raw image to its associated processed outputs and includes anonymized patient-level and image-level descriptors. This metadata file is organized with one row per processed section, and the variables included are summarised in Table \ref{tab:metadata}.
 
\begin{table}[H]
\centering
\caption{Explanation of metadata columns.}
\label{tab:metadata}
\begin{tabular}{lp{0.7\textwidth}}
Column & Description \\ \hline
raw\_image\_path        & {\small Path to the original PNG image stored in \textit{raw/}. Multiple processed images may reference the same raw image.} \\
processed\_image\_path  & {\small Path to the corresponding processed image stored in \textit{processed/}.} \\
patient\_id             & {\small Anonymized numeric identifier assigned to each patient.} \\
image\_type & {\small Image category: S for standard single-frame images, V for images composed of multiple side-by-side frames resulting from vertical image splitting.} \\
sex                     & {\small Patient sex (M/F).}  \\
birth\_date             & {\small Date of birth (DD/MM/YYYY).}  \\
exam\_date              & {\small Date of ERCP examination (DD/MM/YYYY).} \\
exam\_time              & {\small Time of image acquisition (HH:MM:SS).}  \\
age                     & {\small Patient age at the date of examination.}  \\
equipment\_model        & {\small Model name of the imaging equipment used during image acquisition.} \\
Label              & {\small Manually assigned label; may include diagnostic categories or operational labels (e.g., "Unlabelled").} \\
Keep                  & {\small  Binary indicator used to retain or discard specific images during quality review.} \\ \hline
\end{tabular}
\end{table}

\section{Technical Validation}

A fluoroscopic image classifier was developed to automatically distinguish between exams containing lithiasis, bile leaks, strictures and normal findings, based on the available data. Only images with a “Label” column and marked as “Keep” were included. These images were then split into training, validation, and test sets using a stratified strategy based on individual ERCP exams, ensuring that all frames from the same patient were assigned to the same subset. This procedure preserved the class distribution across splits while preventing data leakage. A 70\%–15\%–15\% proportion (train–validation–test) was used.

Multiple deep learning architectures were explored, including MobileNet-V2\cite{sandler2018mobilenetv2}, DenseNet-121\cite{huang2017densenet}, ResNet-50\cite{he2016resnet}, EfficientNet-B7\cite{tan2019efficientnet} and DeiT-III-small\cite{touvron2022deit3}. These architectures were selected to cover a wide range of model complexities and design paradigms. All models were initialized with pretrained weights and subsequently fine-tuned on the ERCP dataset, with the final classification layer replaced to match the number of target classes. Model evaluation followed a hold-out validation approach. Performance was primarily assessed using the F1-score, while accuracy, recall and precision were also examined. EfficientNet-B7 achieved the best overall performance, with a F1-score of 0.609 in the validation set and 0.738 in the test set. Table \ref{tab:bestmodels} presents the performance metrics of each model, based on the test dataset. 

\begin{spacing}{1.3}
\begin{xltabular}{1\textwidth}{@{}%
			>{\centering\arraybackslash}X         
			>{\centering\arraybackslash}m{2cm}   
			>{\centering\arraybackslash}m{2cm}   
			>{\centering\arraybackslash}m{2cm}   
			>{\centering\arraybackslash}m{2cm}     
			>{\centering\arraybackslash}m{2cm}    
			@{}}
		\caption{Performance metrics of each model.} 
		\label{tab:bestmodels}  \\
		\hline
		\multirow{4}{*}{\textbf{Architecture}} & \multicolumn{4}{c}{\textbf{Test}} & \textbf{Validation} \\
		\cmidrule(lr){2-5}
		\cmidrule(lr){5-6}
		                           & \textbf{Accuracy}  & \textbf{Precision (macro)} & \textbf{Recall (macro)} & \textbf{F1-score (macro)}  & \textbf{F1-score (macro)} \\
		\hline
		MobileNetV2              & 61.4                & 58.7                       & 60.5                    & 0.574                      & 0.550 \\
		\hline
		\textbf{EfficientNet-B7} & \textbf{78.3}       & \textbf{75.3}              & \textbf{75.3}           & \textbf{0.738}             & \textbf{0.609} \\
		\hline
		ResNet50                 & 63.7                & 61.7                       & 62.1                    & 0.617                      & 0.490 \\
		\hline
		DenseNet121              & 65.9                & 66.0                       & 63.1                    & 0.621                      & 0.590 \\
		\hline
		DeiT3-Small              & 46.0                & 11.5                       & 25.0                    & 0.158                      & 0.148 \\
		\hline
\end{xltabular}	
\end{spacing} 

\begin{figure}[H]
    \centering
\includegraphics[width=0.7\linewidth]{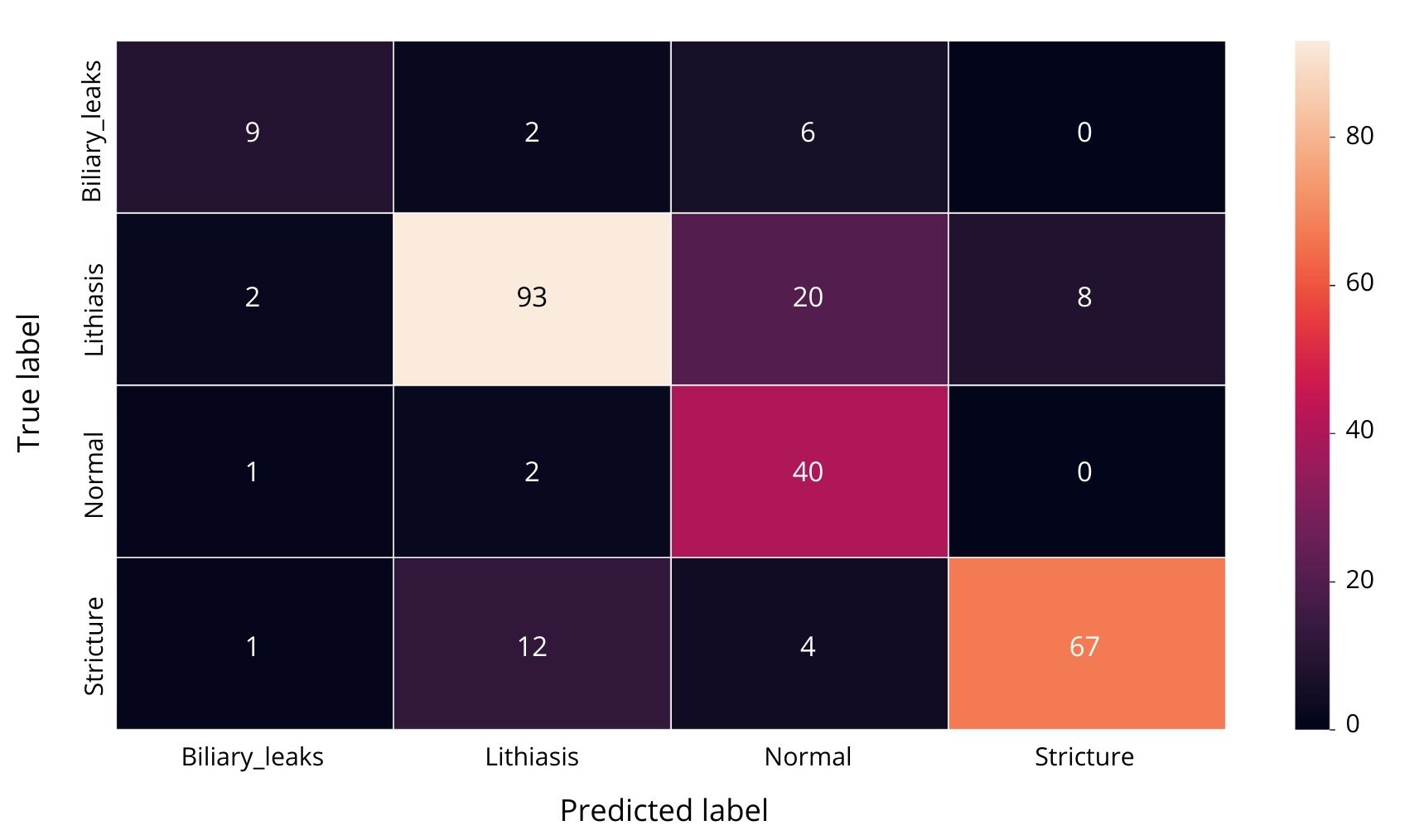}
    \caption{Confusion matrix of the fine-tuned EfficientNet-B7 model evaluated on the test set.}
    \label{fig:confusion_matrix}
\end{figure}

As part of the benchmark evaluation, Figure \ref{fig:confusion_matrix} shows the confusion matrix obtained with EfficientNet-B7, highlighting the separability of the dataset classes.

\section{Usage Notes (optional)}

The dataset is publicly available for download through the Figshare repository available at \url{https://doi.org/10.6084/m9.figshare.31079236}. Due to the dataset size, users are advised to use the direct download link provided in the repository. No special access permissions are required.

All preprocessing steps applied to generate the processed dataset from the raw fluoroscopic images are documented in a public GitHub repository available at: \url{https://github.com/monicaccmartins/MIQR-CC-Dataset}. The repository contains Python scripts for image partitioning and technical validation. These scripts allow reproducibility of the processing pipeline and enable users to adapt or extend the preprocessing workflow to their own data or alternative experimental settings.

\section{Data Availability}
All images were originally acquired using fluoroscopic systems during ERCP procedures. The dataset is openly available in two main folders: "raw" and "processed". The "raw" data contains 19.018 PNG images directly exported from the original DICOM files, with minimal processing applied (only anonymization). The "processed" data comprises 19.317 individual images obtained by partitioning the raw images, each of which may contain multiple sub-images. This preprocessing ensures that the images are ready for direct use in downstream tasks, such as image analysis or machine learning applications, with minimal adaptation. 

A CSV file, \texttt{metadata.csv}, links each raw image to its associated processed images and provides anonymized patient-level and image-level descriptors. The dataset includes labeled cases (\textbf{5.519}) manually inspected and annotated by experts, providing a reliable foundation for benchmarking AI methods in ERCP analysis. Our complete collection is made available at: \url{https://doi.org/10.6084/m9.figshare.31079236}

\section{Code Availability}

All code used for image preprocessing and technical validation is publicly available at:
\url{https://github.com/monicaccmartins/MIQR-CC-Dataset}

\section{Author Contributions}

Alda Andrade performed the export of all fluoroscopic images, conducted image review and annotation, and contributed to data analysis.

Mónica Martins was responsible for image preprocessing, development of the annotation dashboard, data analysis, and co-wrote the manuscript.

Tarcísio Araújo performed ERCP procedures, reviewed and annotated images, and contributed to procedural reports used for cross-checking.

Andre Ferreira contributed to the conceptualization and writing of the manuscript.

Luís Lopes performed ERCP procedures, generated procedural reports, acted as senior reviewer of annotated cases, contributed to data analysis, and co-supervised the study.

Victor Alves conceived the study, provided overall supervision, and secured funding.

All authors contributed to data interpretation, reviewed and approved the final manuscript, and agree to be accountable for their respective contributions.

\section{Competing Interests}

No competing interests.

\section{Funding}

This work has been supported by FCT – Fundação para a Ciência e Tecnologia within the R\&D Unit Project Scope UID/00319/2025 - Centro ALGORITMI (ALGORITMI/UM) \\
https://doi.org/10.54499/UID/00319/2025 and project MIQR-CC with reference 2024.07644.IACDC.

\end{document}